%% file: iclr2027_conference.tex
\PassOptionsToPackage{dvipsnames,table}{xcolor}

\documentclass{article} 
\usepackage{iclr2027_conference,times}

\input{math_commands.tex}

\usepackage{hyperref}
\usepackage{url}

\usepackage[T1]{fontenc}
\usepackage[utf8]{inputenc}
\usepackage{graphicx}
\usepackage{booktabs,tabularx,array,xcolor}
\usepackage{enumitem,needspace}
\usepackage[nopatch=eqnum]{microtype}
\usepackage[section]{placeins}
\newcolumntype{Y}{>{\raggedright\arraybackslash}X}
\definecolor{DraftBlue}{RGB}{35,76,112}
\definecolor{DraftTint}{RGB}{242,246,249}

\usepackage{pifont}
\usepackage{multirow}
\usepackage{fvextra}
\usepackage[breakable,skins]{tcolorbox}

\ifdefined\XeTeXversion
  ="[lmmono10-regular.otf]" at 8pt
  
\fi

\usepackage{wrapfig}
\usepackage{dsfont}

\title{Building AI Research Teams for Autonomous Discovery of Linear Attention Architectures}

\newcommand{\aspace}{\hspace{1.3em}}

\author{Yujian Liu${^{1}}$ \aspace Li An${^{1}}$ \aspace Zelin He${^{2}}$ \aspace Jacob Helwig${^{3}}$ \\
\textbf{Jiabao Ji${^{1}}$ \aspace Rohit Jain${^{3}}$ \aspace Siyu Zhu${^{3}}$ \aspace Shiyu Chang${^{1}}$} \\
$^1$UC Santa Barbara \aspace $^{2}$Pennsylvania State University \aspace $^{3}$LinkedIn \\
\texttt{\{yujianliu,li\_an,jiabaoji,chang87\}@ucsb.edu}
}

\iclrfinalcopy 
\begin{document}

\maketitle

\begin{abstract}
Autonomous agents can increasingly propose ideas, modify code, and run experiments, creating an opportunity to accelerate scientific discovery using teams of collaborative AI researchers. To escape local minima in the design space induced by common reasoning patterns, we propose to encourage orthogonal exploration by constructing the team using agents with varying expertise.
However, realizing this design requires more than diverse backgrounds. When experiments are costly, a team can test only a limited number of ideas, thereby creating the need to balance exploration and exploitation when selecting a subset of researchers to propose and execute new experiments.
Furthermore, each result should shape what the team tries next. The team must decide where to deepen an investigation, when to explore alternatives, and how to carry lessons across researchers.
We therefore ask how AI researchers with different scientific backgrounds can build on one another's findings and sustain progress under a limited experimental budget.
To address these challenges, we build a team of persistent AI researchers, each grounded in a different discipline, who keep private research histories, share experimental evidence, and receive research opportunities through adaptive allocation.
We demonstrate the effectiveness of this design by deploying the team to search for linear attention architectures at 150M parameters. We then scale the team's strongest discovery to 1.3B parameters, where it outperforms GDN2 on language modeling, common-sense reasoning, and synthetic and real-world retrieval tasks.
In controlled comparisons, removing any single organizational component from our design yields markedly weaker discoveries: without disciplinary backgrounds, the search plateaus early; without private histories, researchers make less progress despite identical shared evidence; and uniform selection falls short of adaptive allocation within the same budget.
Tracing the discovery shows researchers applying principles from their own fields to refine one another's ideas. These results suggest that sustained autonomous discovery depends not only on stronger agents, but also on who explores, what they remember, and how research effort is allocated.
\end{abstract}

\input{section/intro}
\input{section/method}
\input{section/experiment}
\input{section/related_work}
\input{section/conclusion}

\subsection*{Acknowledgements}
The UCSB team acknowledges support from the National Science Foundation (NSF) Grant IIS2338252.

\subsection*{AI use statement}
Our use of AI falls into two categories. First, AI-driven research is the subject of this work. The LLM agents in our system propose and refine architectural hypotheses, review proposals, implement methods, interpret experimental results, and share findings. These activities are part of the research methodology described and evaluated in the paper.

Second, we used AI tools to assist the authors in developing and debugging the autoresearch framework, inspecting agent trajectories to diagnose issues in the research workflow, and polishing the manuscript. The authors reviewed these contributions and take responsibility for the final content of the paper and its associated artifacts.

\subsection*{Ethics statement}
This work studies autonomous research in model architecture design using existing training corpora and evaluation benchmarks, without recruiting human participants. The principal ethical concerns involve research integrity: agents may generate incorrect claims, overlook prior work, or optimize evaluation metrics without improving broader capabilities. We retain records of proposals, reviews, implementations, and experimental outcomes to support scrutiny, and use implementation checks and evaluation across multiple capabilities to assess proposed improvements. These safeguards do not guarantee correctness or novelty.

\subsection*{Reproducibility statement}
We document model configurations, training hyperparameters, data preparation, and FLOP-matched budget allocation in Appendix~\ref{app:training}. Appendix~\ref{app:evaluation} specifies the search and full-evaluation protocols, including example selection, decoding settings, scoring rules, and per-dataset results. Appendix~\ref{app:analysis} details researcher initialization, information access, memory conditions, adaptive allocation, and the shared starting point used in organizational comparisons; it also reproduces one complete researcher profile. Appendix~\ref{app:architecture-all} provides the discovered architecture's recurrence, parameterization, and initialization.

\bibliography{iclr2027_conference}
\bibliographystyle{iclr2027_conference}

\clearpage
\appendix
\input{section/appendix_protocol}

\input{section/appendix_architecture}

\end{document}

%% file: math_commands.tex
\usepackage{amsmath,amsfonts,bm}

\def\eqref#1{equation~\ref{#1}}

\def\1{\bm{1}}

\DeclareMathAlphabet{\mathsfit}{\encodingdefault}{\sfdefault}{m}{sl}
\SetMathAlphabet{\mathsfit}{bold}{\encodingdefault}{\sfdefault}{bx}{n}



%% file: section/intro.tex
\section{Introduction}

Autonomous research agents increasingly connect hypothesis generation, implementation, and experimental evaluation into iterative discovery processes \citep{lu2024aiscientistfullyautomated,yamada2025aiscientistv2workshoplevelautomated,schmidgall-etal-2025-agent,novikov2025alphaevolvecodingagentscientific,lange2025shinkaevolveopenendedsampleefficientprogram,karpathy2026autoresearch}.
However, in well-developed fields, progress can enter a local minimum when the explored region of the design space is confined.
Related efforts investigate how persistent experience and collaboration can support scientific work across multiple agents and repeated investigations \citep{qu2026coralautonomousmultiagentevolution,liu2025hypothesishuntingevolvingnetworks,Swanson2024.11.11.623004}.
These capabilities create a compelling opportunity to encourage diverse exploration by assembling a team of AI researchers grounded in different scientific fields, with distinct prior knowledge, conceptual frameworks, and ways of reasoning, and having them work together to push the state-of-the-art on the same open-ended problem. This allows the researchers to leverage a mechanism suggested by one field to address a limitation recognized in another. For example, advances in linear attention have gained insights from fast-weight learning to build memories that can correct stored associations \citep{schlag2021lineartransformerssecretlyfast}. The potential value of such a team therefore lies not only in generating more proposals, but in developing ideas through complementary investigations.

Realizing this potential, however, requires more than assigning researchers different backgrounds. When experiments are costly, a team can test only a limited number of ideas and cannot pursue every direction equally deeply. Which directions will prove productive is uncertain, and due to the high cost of conducting an experiment, exploration and exploitation must be balanced when selecting which of the diverse hypotheses proposed by the team to test.
Furthermore, an early result may reveal a limitation that further work can address. Each experiment therefore does more than test a proposal: it produces evidence that can change what its author or another researcher tries next. The team must use that evidence to decide where to deepen an investigation and when to explore alternatives, while carrying useful lessons across researchers and experiments. Existing demonstrations of collaboration and search have not yet established what disciplinary backgrounds and researchers' own accumulated experience add to shared evidence, or how research opportunities should be distributed across such a team. With these challenges in mind, we therefore ask: \emph{how can a team of AI researchers with different scientific backgrounds build on one another's findings and sustain progress on open-ended research problems under a limited experimental budget?}

\begin{figure}[t]
  \centering
  \includegraphics[width=1\linewidth]{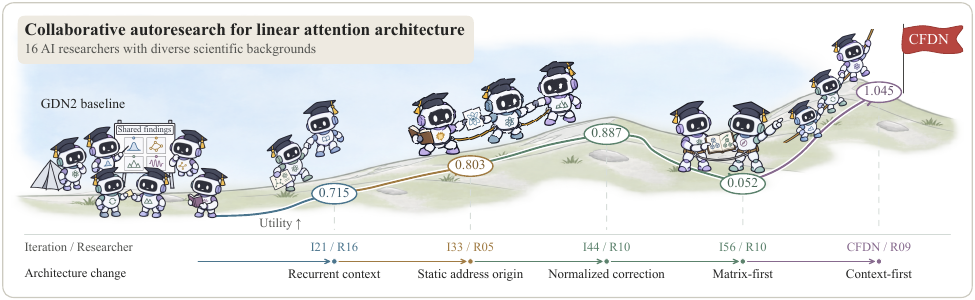}
  \caption{
  \textbf{Illustration of collaborative autoresearch.} 
  Sixteen research agents with diverse scientific backgrounds collaborate on linear attention architecture search. Starting from Gated DeltaNet-2 (GDN2), the team discovers Context-First DeltaNet (CFDN), which improves performance across language modeling, common-sense reasoning, and synthetic and real-world retrieval tasks.
  }
  \label{fig:intro-overview}
\end{figure}

To address these challenges, we propose an autoresearch pipeline that brings together a team of persistent AI researchers, each initialized with a different disciplinary profile containing domain knowledge, mathematical tools, scientific insights, and ways to approach research problems. The team conducts an iterative search in which each iteration follows one proposal through independent review and, if approved, implementation and evaluation, ending with its author reflecting on the outcome.
To make this process cumulative, researchers retain private histories of their investigations while sharing proposals, outcomes, and lessons with the team. This allows individual researchers to develop a line of inquiry over time, while others can revisit its assumptions, extend its mechanisms, or combine its findings with those from another direction. As these lines of inquiry develop, an adaptive scheduler balances further investment in researchers with strong empirical contributions against opportunities for those whose potential is less explored.

To ground this team in a problem that draws on different scientific perspectives, we choose linear attention architecture discovery. Its central modeling challenge is to retain and use sequence information through a fixed-size recurrent state while maintaining model quality and efficient computation \citep{katharopoulos2020transformersrnnsfastautoregressive}. Recent advances have explored connections between state-space models and attention, delta-rule updates, channel-wise forgetting, and separately controlled erasing and writing \citep{dao2024transformersssmsgeneralizedmodels,yang2025parallelizinglineartransformersdelta,yang2025gateddeltanetworksimproving,kimiteam2025kimilinearexpressiveefficient,hatamizadeh2026gateddeltanet2decouplingerase}. These developments illustrate how different perspectives can inform choices about what to remember, how to revise stored information, and how to retrieve it efficiently. Determining which of these choices improve model performance requires implementing and training candidate architectures. The cost of these experiments, in turn, limits how many directions the team can pursue and how deeply it can investigate each one. Building on prior demonstrations that agents can propose, implement, and evaluate architectures \citep{liu2025alphagomomentmodelarchitecture,pepe2026agenticdiscoveryneuralarchitectures}, we therefore use this setting to study how our heterogeneous team organization supports discovery under a limited experimental budget.

We evaluate this organization at 150M model scale, using Gated DeltaNet-2 (GDN2) as the baseline. Proposed candidates receive matched training budgets and are evaluated on language modeling, common-sense reasoning, and retrieval tasks. Our strongest discovery, \emph{Context-First DeltaNet} (CFDN), adds a recurrent vector to GDN2's matrix memory, combining a query-independent contribution from the vector with a query-dependent read from the matrix. We then scale it to 1.3B parameters, where it improves performance in every evaluation category over GDN2.
To understand which organizational choices contribute to these discoveries, we conduct controlled ablations on expert-background conditioning, private longitudinal memory, adaptive researcher selection, and evaluator design. Results show that providing disciplinary profiles broadens the mechanisms explored and leads to better architectures than those discovered without these profiles. Private histories support more effective refinement of earlier ideas beyond what is achieved with shared evidence alone. Adaptive selection finds higher-scoring architectures than uniform selection within the same number of research iterations. Optimizing solely for language modeling, however, steers search toward improvements that transfer less consistently across evaluations.
We complement these comparisons by tracing how ideas develop across researchers. Figure~\ref{fig:intro-overview} provides an overview of the trajectories, which reveal connections between researchers' scientific backgrounds and the hypotheses they pursue, as well as how they adopt and revise each other's ideas using shared experimental evidence.

To summarize, our contributions are threefold:
\begin{itemize}[leftmargin=1.2em,itemsep=2pt,topsep=2pt]
  \item We build a framework for organizing autonomous research teams under limited compute, combining diverse scientific backgrounds, persistent memory, shared evidence, and adaptive allocation.
  \item We discover improved linear attention architectures under matched training compute at 150M scale and evaluate the strongest at 1.3B scale.
  \item We link organizational choices to discovery outcomes through controlled comparisons and trajectory analysis, providing guidance for designing autonomous research teams.
\end{itemize}

%% file: section/method.tex
\section{Methodology}
\label{sec:method}
\begin{figure}[t]
  \centering
  \includegraphics[width=\linewidth]{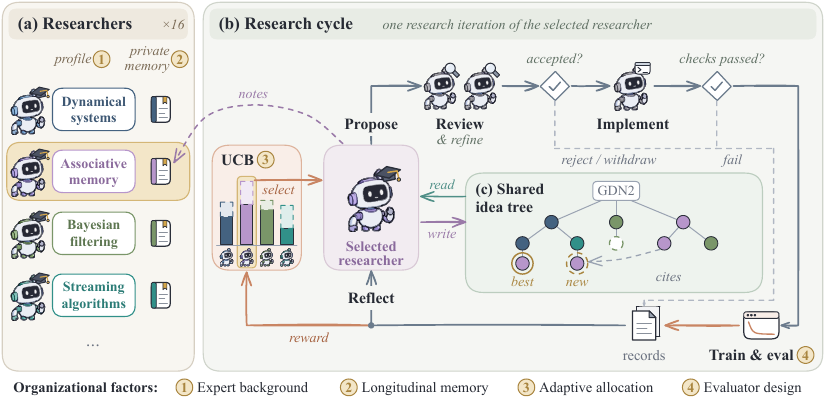}
  \caption{\textbf{Organization of the autonomous research team.} \textbf{(a)} Sixteen researchers have distinct scientific backgrounds and persistent individual histories. Adaptive selection balances measured contributions with opportunities for less-tested researchers. \textbf{(b)} A selected researcher develops one proposal through review and, if approved, implementation and experimentation, then reflects on its outcome. \textbf{(c)} Proposals, outcomes, and their relationships form a shared body of knowledge through which researchers build on one another's work. Circled numbers identify the four organizational factors studied in our comparisons.}
  \label{fig:overview}
\end{figure}

We organize autonomous discovery as a collaboration among researchers who bring different scientific perspectives to a shared problem, encouraging orthogonal exploration of the design space. The team develops through experience, makes each researcher's findings useful to others, and decides where to invest further effort. Figure~\ref{fig:overview} gives an overview of our design. We first describe the researchers and their individual and shared knowledge, then the iterative process through which they develop and test ideas, where costly experiments create a need to balance exploration and exploitation. Finally, we define the experimental feedback that guides our search for linear attention architectures.

\subsection{Scientific Perspectives and Accumulated Knowledge}
\label{sec:team}

In many well-researched areas, substantial progress is hindered by the tendency of researchers to explore confined regions of an otherwise broad space of potential designs, creating a local minimum.
We therefore propose to assemble our AI research team with different disciplinary backgrounds to bring complementary concepts, assumptions, and diagnostic tools to its investigation. Each researcher is an instance of the same language model, with the same tools and a common description of the problem, baseline, and experimental constraints.

\textbf{Grounding researchers in scientific backgrounds.}
Each researcher receives a disciplinary profile that develops three kinds of guidance:
\begin{itemize}[leftmargin=1.2em,itemsep=2pt,topsep=2pt]
  \item \textbf{Prior knowledge and mathematical tools:} central objects, equations, assumptions, and representative literature that provide concrete mechanisms to reason with.
  \item \textbf{Scientific insights:} when those mechanisms are useful, where familiar intuitions fail, and which apparently similar problems require different explanations.
  \item \textbf{Research habits:} ways to decompose an unfamiliar problem, quantities to inspect, and counterexamples that could challenge a hypothesis.
\end{itemize}
For example, the Bayesian-filtering profile explains how prediction errors should be interpreted relative to uncertainty and how to distinguish isolated outliers from persistent changes. A dynamical-systems profile emphasizes how repeated transitions preserve or amplify perturbations. A streaming-algorithms profile asks which queries a bounded summary must support. These perspectives suggest different approaches to the same problem. Appendix~\ref{app:profiles} lists the full roster and example profiles.

\textbf{Preserving individual experience.}
A profile is a starting point for reasoning. Researchers may still develop their perspectives and insights about how their area of expertise relates to the research problem. 
To achieve that, each agent retains its session history and private research notes across investigations, preserving hypotheses, the reasoning behind earlier decisions, and reflections on their outcomes. In addition to persisting these derived insights, this continuity allows a researcher to revisit an unresolved question or revise an assumption.

\textbf{Making findings useful to others.}
Private experience alone would leave useful insights confined to their authors. All researchers therefore have access to a shared record of proposals, reviews, experimental results, and conclusions. It exposes both what was tried and why, allowing researchers to extend a mechanism proposed by others, reconsider an assumption, or combine findings from different investigations. Accumulated findings are synthesized every 10 iterations into shared lessons and unresolved questions. Individual histories and shared knowledge thus support two forms of continuity: developing one's own reasoning and building on discoveries made elsewhere in the team.

\subsection{An Iterative Process for Developing and Testing Ideas}
\label{sec:iteration}

With the initialized researchers, 
the team proposes and evaluates new ideas to approach the problem in an iterative process. Each iteration contains the following steps that turn existing knowledge into a new investigation and return its findings to the team.

\textbf{Step 1: Select a researcher.}
Although we would ideally allow all researchers to propose and test diverse ideas in each iteration, this is not tractable in most real-world research settings due to high experimental costs. Therefore, in each iteration, we must first select one researcher from the team to propose a new idea. With limited experimental opportunities, repeatedly choosing recent winners can overlook researchers whose ideas have not been sufficiently tested. We therefore use discounted upper confidence bound (UCB) selection~\citep{10.1007/978-3-642-24412-4_16} to balance exploration and exploitation.
Each completed iteration $j$ earns a nonnegative reward $r_j$ for improving on the team's best model known when that iteration began. Section~\ref{sec:progress} defines this reward in detail.

Specifically, let $t$ count completed iterations, $t_j$ be iteration $j$'s completion ordinal, and $\mathcal{J}_i$ index the $i$-th researcher's completed iterations. For researcher $i$, we use discount factor $\gamma \in (0, 1)$ to compute an effective number of observations $N_i$, mean reward $\bar r_i$, and selection score $A_i$ as 
\begin{equation}
  N_i(t)=\sum_{j\in\mathcal{J}_i}\gamma^{t-t_j},\quad
  \bar r_i(t)=\frac{\sum_{j\in\mathcal{J}_i}\gamma^{t-t_j}r_j}{N_i(t)},\quad
  A_i(t)=\bar r_i(t)+c\sqrt{\frac{\log(\sum_k N_k(t))}{N_i(t)}}.
\label{eq:ucb}
\end{equation}
The researcher with the highest score $A_i(t)$ is selected for the next iteration.
The mean reward $\bar r_i(t)$ favors researchers with stronger historical contributions, and the second term preserves opportunities for less-tested researchers, with $c$ being a hyperparameter controlling the strength of exploration. Intuitively, the discount factor $\gamma$ assigns more weight to recent contributions, thus giving researchers who perform well recently more continuity in developing along their promising directions.

\textbf{Step 2: Propose a hypothesis.}
Selection determines who investigates next, but the researcher decides which scientific direction to pursue. The selected researcher examines its own history and shared experimental results to identify a limitation worth investigating. It may develop a new mechanism or revise an earlier proposal, including one from another researcher. The proposal states what should change, why that change should help, and which experimental outcome would challenge the hypothesis. For our linear attention architecture discovery, it specifies the mathematical operations and their parameterization, together with a parallelizable formulation suitable for training.

\textbf{Step 3: Review the proposal.}
Before committing training compute, two independent reviewers assess whether the proposed experiment is sound and informative. A technical reviewer checks the mathematics and computational feasibility. If these checks pass, a second reviewer assesses novelty and research value in light of prior work and the team's findings. A rejection leads directly to reflection (Step~6); otherwise, the reviews identify any issues to resolve before testing the proposal.

\textbf{Step 4: Refine the proposal.}
For a proposal that passes review, the author revises its mathematical design in response to the critiques. The author is responsible for resolving the reviewers' critiques and providing a detailed specification for implementation. If the critiques cannot be addressed without replacing the central mechanism, the author withdraws the proposal and proceeds to reflection.

\textbf{Step 5: Implement, train, and evaluate.}
A separate implementation agent translates the final specification into code and checks that it realizes the intended computation. The implementation must pass integration and speed checks before training. Specifically, we require the implemented model's per-step training time to be less than 1.2 times the baseline GDN2 time to prevent architectures that degrade efficiency. Repairable failures are returned to the implementer with diagnostics.
Finally, validated candidates are trained from scratch under matched training compute, using the same training data, and evaluated with three metrics defined in Section~\ref{sec:progress}.

\textbf{Step 6: Reflect and share.}
The proposing researcher revisits its original hypothesis in light of the outcome, including rejection or failure. Reviews, implementation, and experimental records are provided to help it distinguish evidence about the mechanism from problems in realizing or testing it. The researcher records what the outcome supports, what remains unresolved, and what it suggests investigating next. It updates its private notes for future reference and shares generalizable lessons with other researchers. Completing this reflection ends the iteration and updates the observations used for researcher selection in future iterations.

\subsection{Evaluating Architectures and Rewarding Research Progress}
\label{sec:progress}

The evaluation metrics shape both the hypotheses researchers pursue and the reward used by the selection rule. For linear attention discovery, we seek improvements in general model capabilities that span different tasks and scenarios. We therefore evaluate three complementary aspects to measure progress on model quality (details in Appendix~\ref{app:evaluation}).

\ding{182} \textbf{Language modeling} measures next-token prediction with negative log-likelihood $L$; lower is better. \ding{183} \textbf{Common-sense reasoning} measures downstream task performance with mean accuracy $Q$; higher is better. \ding{184} \textbf{Retrieval} tests whether information remains accessible from context, using an exact-continuation proxy $R$ on the synthetic needle-in-a-haystack tasks; higher is better.

\textbf{Model utility.}
To assess the utility of a specific trained model, we measure its metrics $(L_x,Q_x,R_x)$ and calculate its normalized improvements over the baseline GDN2's metrics $(L_0,Q_0,R_0)$:
\begin{equation}
  U(x)=\frac{1}{3}\left(
  \frac{L_0-L_x}{s_L}
  +\frac{Q_x-Q_0}{s_Q}
  +\frac{R_x-R_0}{s_R}\right),
  \label{eq:utility}
\end{equation}
where $(s_L,s_Q,s_R)$ are scales informed by metric variability in preliminary experiments and remain fixed throughout all experiments. Researchers see both the aggregated utility and individual metrics, allowing them to inspect tradeoffs among the metrics.

\textbf{Researcher reward.}
A high model utility may not indicate a new contribution if a better model was already available. At the start of iteration $j$, we therefore record the architecture with the highest utility, including the baseline, as $x_j^{\mathrm{ref}}$. Candidate architecture $x_j$ proposed in iteration $j$ then earns reward $r_j=\max\{0,\,U(x_j)-U(x_j^{\mathrm{ref}})\}$, which supplies the reward used for selection. Rejected proposals or implementation failures receive zero reward. Intuitively, this definition rewards overall improvements compared to the best candidate so far, without assigning negative rewards to unsuccessful attempts, thus encouraging researchers to try out promising directions.

%% file: section/experiment.tex
\section{Experiments}
\label{sec:experiments}

Our experiments answer three questions: can the team discover useful architectures; which organizational choices help in the limited-iteration setting; and how do different researchers contribute to a successful discovery? We first describe the experiment settings, then evaluate the discovered architecture, compare research organizations, and trace the development of the found architecture.

\subsection{Experimental Setting}
\label{sec:setup}

\textbf{Research team.}
We use 16 researchers and a budget of 80 research iterations. All agent roles use GPT-6 Astra with high reasoning effort through the Codex harness. We allow a maximum of 8 researchers working in parallel, while forcing their proposal stage to be sequential to prevent repetition in proposed ideas. In the first 16 iterations, all researchers are invoked exactly once, and further researcher selection is based on the UCB rule.

\textbf{Model backbones and search space.}
We search at approximately 150M non-embedding parameters, and we scale the discovered best candidate to 1.3B parameters. We use GDN2~\citep{hatamizadeh2026gateddeltanet2decouplingerase} as the baseline. Candidates can change the recurrent mixer's memory representation, update rule, readout, and parameterization, while retaining the common depth, residual width, and feed-forward network. We also compare with Gated DeltaNet (GDN)~\citep{yang2025gateddeltanetworksimproving} and Kimi Delta Attention (KDA)~\citep{kimiteam2025kimilinearexpressiveefficient} using the same surrounding architecture. Appendix~\ref{app:training} gives the dimensions and parameter-count convention at both scales.

\textbf{Training data and budget.}
All models are trained from scratch on FineWeb-Edu~\citep{penedo2024finewebdatasetsdecantingweb}. For all models, we adjust their training tokens to match the total FLOPs of training GDN: at 150M scale, we train GDN for 15B tokens; at 1.3B scale, we train GDN for 100B tokens. More expensive architectures therefore receive fewer training tokens. Following \citet{hatamizadeh2026gateddeltanet2decouplingerase}, all models use the same optimizer hyperparameters and global batch size (details in Appendix~\ref{app:training}).

\textbf{Final evaluation and search feedback.}
We evaluate models on language modeling (LM), common-sense reasoning (CSR), synthetic needle-in-a-haystack (NIAH), and, for 1.3B models, in-context real-world retrieval (RWR) tasks. For LM, we report perplexity on WikiText (Wiki.)~\citep{merity2016pointersentinelmixturemodels} and LAMBADA (LMB.)~\citep{paperno-etal-2016-lambada}. For CSR, we report the mean accuracy over eight datasets~\citep{bisk2019piqareasoningphysicalcommonsense,zellers-etal-2019-hellaswag,sakaguchi2019winograndeadversarialwinogradschema,clark2018thinksolvedquestionanswering,mihaylov2018suitarmorconductelectricity,sap2019socialiqacommonsensereasoningsocial,clark2019boolqexploringsurprisingdifficulty}. For NIAH, we use four synthetic tasks (S-NIAH-1, 2, 3 and MK-NIAH) from RULER~\citep{hsieh2024rulerwhatsrealcontext}. For in-context real-world retrieval, we report mean accuracy over six datasets~\citep{arora2025simplelinearattentionlanguage,rajpurkar-etal-2018-know,joshi2017triviaqalargescaledistantly,kwiatkowski-etal-2019-natural,dua2019dropreadingcomprehensionbenchmark}.

During search, we use a reduced suite to provide feedback: LM loss on the same two datasets, a subset of 256 examples from each CSR dataset, and 64 examples per NIAH task at each of 1K, 2K, and 4K context. We do not evaluate RWR during search. The NIAH evaluation uses a teacher-forced continuation proxy instead of answer generation. Appendix~\ref{app:evaluation} gives more details on evaluation.

\subsection{Discovered Architecture and Validation at Larger Scale}
\label{sec:discovery}
The search run contains 63 candidate evaluations in 80 iterations.\footnote{Rejected or failed proposals also count toward the budget, so it is not a guarantee of 80 trained models.} We select Context-First DeltaNet (CFDN), the candidate with the highest search utility, and compare it with GDN, KDA, and GDN2.

\begin{wraptable}{r}{0.55\textwidth}
\vspace{-2em}
\centering
\caption{\textbf{150M comparison.} $U$ is the aggregated search utility; CSR is the mean of eight common-sense reasoning tasks; NIAH is the mean of four synthetic retrieval tasks. Full results are in Appendix~\ref{app:evaluation}.}
\label{tab:search}
\resizebox{\linewidth}{!}{%
\begin{tabular}{lcccccc}
\toprule \midrule
& & \textbf{Search} & \multicolumn{4}{c}{\textbf{Full evaluation}} \\
\cmidrule(lr){3-3}\cmidrule(l){4-7}
\multirow{2}{*}{\textbf{Model}} & \textbf{Tokens} & \textbf{Utility} & \textbf{Wiki.} & \textbf{LMB.} & \textbf{CSR} & \textbf{NIAH} \\
& (B) & $U \uparrow$ & ppl $\downarrow$ & ppl $\downarrow$ & acc $\uparrow$ & acc $\uparrow$\\
\midrule
GDN & 15.00 & -1.29 & 35.15 & 64.03 & 42.88 & 16.94\\
KDA & 14.82 & -0.59 & 34.63 & 56.64 & 41.94 & 22.03\\
GDN2 & 13.53 & 0.00 & 34.93 & 54.74 & \textbf{43.60} & 25.79\\
\rowcolor{gray!20}
CFDN & 13.53 & \textbf{1.04} & \textbf{33.84} & \textbf{52.33} & 43.29 & \textbf{29.76}\\
\midrule \bottomrule
\end{tabular}
}
\end{wraptable}

\textbf{What was discovered?}
CFDN augments GDN2's recurrent matrix memory with a recurrent vector. The vector contributes the same context information to every query, while the matrix supplies a query-dependent prediction. Given the current token's key, the memories jointly predict its value. The defining change is how they learn from that prediction error: the vector is updated first, and the matrix then fits the remaining error. This gives shared information priority over query-specific correction. The resulting update does not add notable compute over GDN2 and remains compatible with efficient parallel training. Section~\ref{sec:lineage} explains how this mechanism is developed across researchers; Appendix~\ref{app:architecture} gives the full recurrence.

As shown in Table~\ref{tab:search}, despite using the fewest training tokens, CFDN achieves a substantially higher utility than all baselines. Specifically, it improves all three search targets over GDN2: LM perplexity decreases from 12.21 to 11.90, CSR accuracy rises from 44.48\% to 44.63\%, and retrieval rises from 28.91\% to 50.00\%.
In terms of the final full evaluation, CFDN outperforms all baselines on two language modeling tasks, and more considerably on the NIAH tasks. Its CSR performance is slightly worse than GDN2 but still better than GDN and KDA.
These results demonstrate that \textbf{the search found a genuinely better architecture than the baselines}.

\begin{wraptable}{r}{0.55\textwidth}
\vspace{-2em}
\centering
\caption{\textbf{Full evaluation at 1.3B.} CSR is the mean of eight common-sense reasoning tasks; NIAH is the mean of four synthetic retrieval tasks; RWR is the mean of six in-context real-world retrieval tasks. Full results are in Appendix~\ref{app:evaluation}.}
\label{tab:large}
\resizebox{\linewidth}{!}{%
\begin{tabular}{lcccccc}
\toprule \midrule
\multirow{2}{*}{\textbf{Model}} & \textbf{Tokens} & \textbf{Wiki.} & \textbf{LMB.} & \textbf{CSR} & \textbf{NIAH} & \textbf{RWR} \\
& (B) & ppl $\downarrow$ & ppl $\downarrow$ & acc $\uparrow$ & acc $\uparrow$ & acc $\uparrow$\\
\midrule
GDN & 100.00 & 16.27 & 12.04 & 53.73 & 54.99 & 33.17\\
KDA & 99.24 & 16.05 & 11.66 & 53.90 & 59.11 & 32.47\\
GDN2 & 87.18 & 16.05 & 11.22 & 52.65 & 66.86 & 32.52\\
\rowcolor{gray!20}
CFDN & 87.17 & \textbf{15.74} & \textbf{11.04} & \textbf{54.53} & \textbf{67.29} & \textbf{33.41}\\
\midrule \bottomrule
\end{tabular}
}
\end{wraptable}

\textbf{Validation at 1.3B.}
To validate whether the discovered architecture remains effective at a larger scale, we train CFDN and all baselines at around 1.3B parameters. Table~\ref{tab:large} compares the same four architectures under the GDN-100B reference budget, using the full evaluation protocol.

At 1.3B, CFDN achieves the best performance in each evaluation category among these models. It obtains the lowest perplexity on both LM datasets and the highest mean CSR, NIAH, and real-world retrieval accuracy. This validates that \textbf{the discovered architecture remains effective at a larger scale than the search region}.

\subsection{Which Organizational Choices Support Discovery?}
\label{sec:ablations}

A useful architecture does not by itself explain whether the team's organization helps discovery. We therefore compare the full system with three variants, holding the agent model, iteration allowance, per-candidate training budget, and search feedback fixed. \ding{182} \emph{No background} removes disciplinary profiles while retaining researcher identities and their private histories. \ding{183} \emph{Fresh memory} removes the researchers' private histories and notes between iterations, while preserving their disciplinary profiles. \ding{184} \emph{Uniform selection} replaces the UCB-defined adaptive selection with a uniform selection that repeats researchers in a fixed order. All variants retain access to shared proposals, experimental results, and lessons, so each comparison concerns a design beyond this common evidence.

The main and no-background runs start from scratch. To reduce variation due to different initial discoveries, fresh-memory and uniform-selection both reuse the main run's first 16 iterations (one iteration per researcher) and conduct 64 further iterations. Appendix~\ref{app:analysis} gives more details.

\begin{wrapfigure}{r}{0.55\textwidth}
\vspace{-3pt}
\centering
\includegraphics[width=\linewidth]{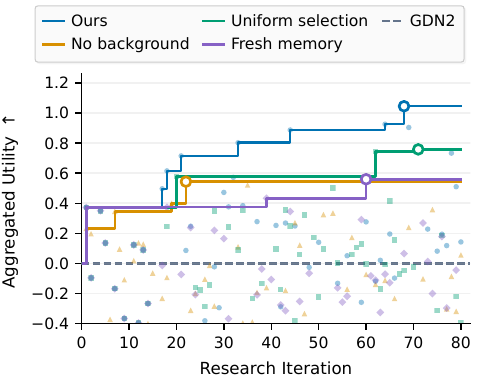}
\vspace{-2em}
\caption{\textbf{Candidate utility throughout the search.} Points are evaluated candidates; lines are the best found candidates up to an iteration.}
\label{fig:organization}
\vspace{-5pt}
\end{wrapfigure}

Figure~\ref{fig:organization} shows the utility of each candidate in the search process. The full system discovers the strongest architecture in these comparisons: its best utility is 1.04, versus 0.54 without backgrounds, 0.56 with fresh memory, and 0.76 with uniform selection. The difference is also not explained by completing more model evaluations: the main run contains 63 evaluated candidates, compared with 65-68 in the variants.

Interestingly, the main run's strongest proposal appears at iteration 68, whereas the no-background run reaches its best utility at iteration 22 and does not improve it thereafter, which suggests that \textbf{the diverse scientific backgrounds help escape the local minima}. The full system also reaches higher utility than the fresh-memory and uniform-selection runs, indicating that \textbf{the private researcher memory and adaptive allocation facilitate the team to make sustained progress} on the problem.

\subsection{How Does the Search Objective Steer Discoveries?}
\label{sec:evaluator}

We next study how the choice of search objective affects the architectures discovered. We compare the three-target objective with an independent LM-only search, retaining the same researcher profiles, private memory, adaptive selection rule, and experimental budget. The LM-only condition uses the same WikiText and LAMBADA perplexity but excludes CSR and NIAH from the objective. Each condition selects its strongest candidate using its own objective.

\begin{wraptable}{r}{0.55\textwidth}
\vspace{-2em}
\centering
\caption{\textbf{Effect of the search objective.} Full evaluation of the model discovered at 150M scale.}
\label{tab:objective}
\resizebox{\linewidth}{!}{%
\begin{tabular}{lcccc}
\toprule \midrule
\multirow{2}{*}{\textbf{Search objective}} & \textbf{Wiki.} & \textbf{LMB.} & \textbf{CSR} & \textbf{NIAH}\\
& ppl $\downarrow$ & ppl $\downarrow$ & acc $\uparrow$ & acc $\uparrow$\\
\midrule
LM only & 34.59 & \textbf{46.40} & 42.25 & 26.73\\
LM $+$ CSR $+$ NIAH & \textbf{33.84} & 52.33 & \textbf{43.29} & \textbf{29.76}\\
\midrule \bottomrule
\end{tabular}
}
\end{wraptable}

Table~\ref{tab:objective} compares the full evaluation of two models at 150M scale. As can be observed, LM-only search achieves a substantially lower perplexity on LAMBADA, while three-target search yields stronger results on the other metrics.
These results indicate that objective choice matters for discovery: gains under an LM-only objective \textbf{may not extend to all capabilities of interest}. This is relevant to popular autoresearch systems guided by LM quality, such as~\citet{karpathy2026autoresearch}, which uses validation bits per byte as its objective. Although perplexity captures broad predictive performance, \textit{averaging over many token predictions can dilute the influence of the few answer tokens that determine success in the tasks of interest}, such as retrieval and common-sense reasoning, motivating explicit evaluation of these capabilities.

\subsection{How Researchers Develop the Discovered Architecture}
\label{sec:lineage}

Figure~\ref{fig:intro-overview} traces CFDN's development through contributions from four researchers. A researcher grounded in linear attention introduces a recurrent context vector alongside GDN2's matrix memory, building on affine-memory precedent~\citep{zuo2026parallaxparameterizedlocallinear}. Another researcher learns the address origin that determines what this vector represents. A researcher with expertise in adaptive filtering then normalizes the correction by the translated address's energy, showing how \textbf{disciplinary knowledge can refine a colleague's mechanism}.

The next revision updates the matrix before the context vector, improving LM perplexity but weakening CSR and retrieval. A researcher with a background in predictive representations uses this tradeoff to reconsider which memory should absorb the prediction error first. It then switches to updating shared context before the matrix, which produces CFDN. Thus, the team uses a degraded result to guide a subsequent improvement, rather than simply retaining successful proposals.

%% file: section/related_work.tex
\section{Related Work}

\textbf{Autonomous research agents.}
A growing line of work automates the iterative research process from hypothesis to experiment, through fixed pipelines from idea generation to written paper~\citep{lu2024aiscientistfullyautomated, schmidgall-etal-2025-agent}, agentic tree search~\citep{yamada2025aiscientistv2workshoplevelautomated}, or evolutionary search with the language model as the mutation operator~\citep{novikov2025alphaevolvecodingagentscientific, lange2025shinkaevolveopenendedsampleefficientprogram}.
In these works, a single agent or a fixed pipeline proposes every candidate.

Other work organizes a team of researchers around a shared scientific problem.
These teams differ in how they are composed, what each member retains and what the team shares, and how research opportunities are distributed.
VirSci, ASCollab, and ScienceClaw~\citep{su-etal-2025-many, liu2025hypothesishuntingevolvingnetworks, wang2026autonomousagentscoordinatingdistributed} give agents distinct backgrounds, personas, or scientific profiles.
CORAL~\citep{qu2026coralautonomousmultiagentevolution}, ASCollab, and ScienceClaw keep long-running agents with their own histories and a shared record, whereas AgentRxiv~\citep{schmidgall2025agentrxivcollaborativeautonomousresearch} starts a fresh laboratory instance per project that reads a shared archive of earlier papers.
However, none compares teams with and without different backgrounds on the outcomes of implementing, training, and evaluating what they propose, or tests whether private memory adds anything once findings are shared.

Adaptive allocation itself is not new, but no existing work both gives researchers different backgrounds and allocates opportunities among them adaptively.
The Virtual Lab~\citep{Swanson2024.11.11.623004} decides which specialist takes an agenda item, ShinkaEvolve~\citep{lange2025shinkaevolveopenendedsampleefficientprogram} decides which language model performs the next mutation, SwarmResearch~\citep{virk2026swarmresearchorchestratingcodingagents} decides which branch of the search to extend, and Economy of Minds~\citep{qi2026economymindsemergingmultiagent} decides which agent acts next in a task.
Each of them chooses an operator or a candidate to advance.
In our setting, the decision is which researcher proposes and tests the next idea, and because each researcher brings its own background and record of what it has tried and learned, that choice determines what evidence the team will have next.
No prior team of researchers with different backgrounds makes this decision adaptively.

Autonomous research has also been applied to architecture design.
NADER~\citep{yang2024naderneuralarchitecturedesign} coordinates fixed-role agents that propose changes to a tree of prior modifications. AIRA-Compose and AIRA-Design~\citep{pepe2026agenticdiscoveryneuralarchitectures} pool designs from independent search agents, and ASI-Arch~\citep{liu2025alphagomomentmodelarchitecture} runs a researcher, engineer, and analyst pipeline over linear attention architectures and reports that discoveries grow with compute.
These systems use fixed-role workflows or independent agent searches; we instead fix the experimental budget and vary the organization, comparing who proposes, what they retain and share, who receives the next opportunity, and what counts as progress.

\textbf{Linear attention.}
Linear attention replaces the key-value cache of softmax attention with a fixed-size recurrent state~\citep{katharopoulos2020transformersrnnsfastautoregressive}, and subsequent designs mainly differ in what the state remembers, how stored information is revised, and how it is retrieved.
On what to remember, most designs keep a matrix of key–value associations, while Titans~\citep{behrouz2024titanslearningmemorizetest} replaces it with a network learned at inference.
On how to revise, RetNet~\citep{sun2023retentivenetworksuccessortransformer} decays the state at a fixed rate, GLA and Mamba-2~\citep{yang2024gatedlinearattentiontransformers, dao2024transformersssmsgeneralizedmodels} make the decay input-dependent, DeltaNet~\citep{yang2025parallelizinglineartransformersdelta} replaces additive writes with delta-rule correction, GDN and KDA~\citep{yang2025gateddeltanetworksimproving, kimiteam2025kimilinearexpressiveefficient} combine the delta rule with learned and fine-grained gating, and GDN2~\citep{hatamizadeh2026gateddeltanet2decouplingerase} separates erasing from writing with channel-wise gates.
On how to retrieve, MesaNet~\citep{vonoswald2026mesanetsequencemodelinglocally} solves the regression to optimality at every step, and Parallax~\citep{zuo2026parallaxparameterizedlocallinear} restores the intercept of the underlying regression.
Each of these works changes one or two choices along these axes, and the space they leave open is large, which makes linear attention architecture discovery a natural problem for an autonomous research team.

%% file: section/conclusion.tex
\section{Conclusion}

We study how to organize autonomous researchers for cumulative discovery under limited experimental budgets. Our framework combines diverse scientific backgrounds, persistent memory, shared evidence, and adaptive allocation of research opportunities. Applied to linear attention, the team discovered CFDN, which improves over GDN2 at two scales. Organizational comparisons and research trajectories highlight the importance of heterogeneous agents, adaptive allocation, and shared experimental records. Together, these results suggest that sustained autonomous discovery depends on how a team learns from experiments and decides which ideas to pursue next.

%% file: section/appendix_protocol.tex
\section{Training and compute protocol}
\label{app:training}

Tables~\ref{tab:training} and~\ref{tab:training-hyperparameters} summarize the model configurations and training hyperparameters, respectively.

\begin{table}[!htbp]
\centering
\caption{\textbf{Model configurations.} Recurrent-head geometry describes GDN2; the outer backbone is fixed within each scale. Scale labels exclude the input embedding and output vocabulary projection. Both remain trainable, and the output projection is included in reference FLOPs.}
\label{tab:training}
\small
\begin{tabular}{@{}lcc@{}}
\toprule \midrule
\textbf{Setting} & \textbf{Search scale} & \textbf{Large-scale comparison}\\
\midrule
Blocks & 8 & 18\\
Residual / FFN width & 1,024 / 4,096 & 2,304 / 6,208\\
GDN2 recurrent heads & $8\times128$ & $16\times128$\\
GDN2 non-embedding parameters & 155,322,432 & 1,302,640,416\\
GDN2 total parameters & 220,858,432 & 1,450,096,416\\
Output gate & Sigmoid & Sigmoid\\
Vocabulary size & 32,000 & 32,000\\
\midrule \bottomrule
\end{tabular}
\end{table}

\begin{table}[!htbp]
\centering
\caption{\textbf{Training hyperparameters.} The optimizer settings and global batch size are shared by all architectures. Microbatch size denotes the sequences processed per device in one forward/backward pass.}
\label{tab:training-hyperparameters}
\small
\setlength{\tabcolsep}{5pt}
\begin{tabular}{@{}lcc@{}}
\toprule \midrule
\textbf{Hyperparameter} & \textbf{Search scale} & \textbf{1.3B scale}\\
\midrule
Optimizer & AdamW & AdamW\\
Peak learning rate & $4\times10^{-4}$ & $4\times10^{-4}$\\
Minimum learning rate & $4\times10^{-5}$ & $4\times10^{-5}$\\
Adam $\beta_1,\beta_2$ & 0.9, 0.95 & 0.9, 0.95\\
Weight decay & 0.1 & 0.1\\
Gradient clipping norm & 1.0 & 1.0\\
Warmup fraction & 1\% & 1\%\\
Learning-rate decay & Cosine & Cosine\\
Context length & 4,096 & 4,096\\
Global batch (sequences) & 512 & 512\\
Tokens per optimizer update & 2,097,152 & 2,097,152\\
Microbatch (sequences) & 8 & 4\\
Accumulation steps & 8 & 2\\
Training seed & 3407 & 3407\\
Compute anchor (GDN tokens) & 15B & 100B\\
\midrule \bottomrule
\end{tabular}
\end{table}

\textbf{Common small-scale data stream.}
The search corpus is prepared once from 24 FineWeb-Edu source shards with the TinyLlama tokenizer. Within each shard, tokenized documents are concatenated and packed into 4,097-ID blocks, yielding 4,096 input--target positions per sequence. Long documents span multiple blocks; only each shard's final incomplete block is discarded. One saved permutation determines the global sequence order. Each candidate consumes the first $T(x)/4096$ sequences of that same order, without reshuffling after budget selection. An optimizer update advances by 512 sequences.

\textbf{Baseline parameterization.}
GDN2 uses separate full projections for queries, keys, values, and channel-wise erase and write gates. Query, key, and value projections are followed by causal depthwise convolutions of width four and SiLU; queries and keys are L2-normalized. Decay and output-gate preactivations use two linear maps with bottleneck width 128. The recurrent read is RMS-normalized per head, multiplied by a sigmoid output gate, and projected back to the residual stream. Our KDA comparison shares these surrounding components and initialization but uses one scalar erase/write gate per head; GDN additionally uses scalar rather than channel-wise state decay. The architectural differences are therefore confined to the recurrent mechanisms within a common surrounding backbone, rather than unrelated released model recipes.

\subsection{Architectural reference FLOPs and practical constraints}
\label{app:compute}

For training-compute budget $\mathcal C$, reference FLOPs per token $F(x)$, and $B$ processed token positions per optimizer update, search allocates $T(x)=B\lfloor\mathcal C/(B F(x))\rfloor$ tokens. Rounding down to complete updates keeps each candidate within the nominal budget.

Let $P_{\mathrm{proj}}$ count active dense and convolutional weights, and let $C_{\mathrm{mem}}$ and $C_{\mathrm{extra}}$ denote forward reference arithmetic per token for memory computation and any other substantial tensor operations. The allocation uses
\begin{equation}
 F(x)=6P_{\mathrm{proj}}+3\bigl(C_{\mathrm{mem}}+C_{\mathrm{extra}}\bigr).
 \label{eq:flops}
\end{equation}
Input-embedding lookup is excluded; the language-model output projection is included. Biases, normalization scales, and static parameters are inventoried separately from projection weights. Repeated uses of a shared projection are charged for their additional computation. For custom operators, the forward count is multiplied by three as a uniform training-cost approximation.

The general counting rule is to include projections and memory computation, while excluding ordinary token-local scalar/vector processing. Memory reads, contractions, outer-product updates, and transformations of stored state are counted. Ordinary activations, normalization, scalar/vector gates, residual additions, embedding lookup, loss computation, optimizer steps, communication, and checkpointing are excluded. This is a consistent architectural proxy, not a hardware instruction count.

The proposer derives the breakdown, the technical reviewer checks it, and the author's revision resolves disagreements. Automated checks measure the implemented weight inventory. A mismatch receives one implementer recheck before the measured counts determine the projection term; the custom-operation terms remain those approved in scientific review. Neither parameter counting nor finite training alone proves that an arbitrary implementation faithfully realizes the specification.

For the small GDN compute anchor, $F=1{,}037{,}893{,}632$ FLOPs/token, giving $\mathcal{C}=1.556840448\times10^{19}$ FLOPs. GDN2 costs $1{,}150{,}353{,}408$ FLOPs/token and receives 6,453 updates; CFDN costs $1{,}150{,}500{,}864$ FLOPs/token and receives 6,452.

Practical speed is checked separately. Candidate optimizer-step time must be at most $1.2\times$ the baseline measured on the same hardware. During training, the candidate must reach 25\%, 50\%, 75\%, and 100\% of its own allocation within $1.2\times$ the corresponding baseline elapsed times. These checks can reject an implementation even when its analytical compute is small. Ordinary implementation repairs receive at most three rounds of validation and repair and six active hours; infrastructure failures do not count as scientific failures. These constraints define the implemented search space and must be retained in organizational comparisons.

\section{Evaluation definitions and complete result tables}
\label{app:evaluation}

\textbf{Search suite.}
WikiText-2 (Wiki.) uses 62 documents scored through independent rolling 4K windows, totaling 322,149 scored tokens. LAMBADA (LMB.) uses 5,153 examples and scores their final-word continuations, totaling 8,808 tokens. We average the two token-mean negative log-likelihoods equally before exponentiation.

CSR uses 256 fixed examples from each of PIQA, HellaSwag, WinoGrande, ARC-Easy, ARC-Challenge, OpenBookQA, Social IQa, and BoolQ. Option likelihoods follow the frozen zero-shot requests. HellaSwag uses character-normalized accuracy; the other tasks use raw accuracy. The reported $Q$ is the unweighted mean over these eight tasks.

Retrieval uses S-NIAH-1, S-NIAH-2, S-NIAH-3, and MK-NIAH-1, with 64 fixed examples at each of 1K, 2K, and 4K, totaling 768. The evaluator reveals the first continuation token and teacher-forces the rest. An example passes only if every remaining answer token is the model's argmax. The first token is not scored. The reported $R$ averages the 12 task/length cells equally. This proxy does not require autoregressive generation.

\textbf{Full evaluation.}
The full protocol uses all examples in the selected LM/CSR datasets and generated-answer NIAH scoring. NIAH contains 500 frozen requests per task/length cell: S-NIAH-1/2 at 1K, 2K, 4K, and 8K, and S-NIAH-3/MK-NIAH-1 at 1K, 2K, and 4K, for 7,000 requests in 14 cells. All models receive the same requests. Generation is greedy, with at most 128 new tokens and early stopping at EOS. We use batch size eight, BF16, and cached decoding. The RULER scorer checks whether each reference answer appears in the generated text as a case-insensitive substring and averages the fraction of references recovered per request; it does not require the entire generated string to equal the answer. The NIAH mean weights all cells equally.

\textbf{Real-world retrieval at 1.3B.}
We use the released BASED/JRT cloze-completion suite~\citep{arora2025simplelinearattentionlanguage}: SWDE, SQuAD, FDA, TriviaQA, Natural Questions (NQ), and DROP. This suite is used only for final evaluation of the 1.3B models and does not enter search feedback or utility. The nominal context length is 2K. We retain the released preprocessing and its exclusions for empty or single-character primary answers, leaving 12,077 examples: 1,067, 2,977, 1,101, 1,688, 3,157, and 2,087 in the dataset order above. Context is cropped before appending the cloze query; two final NQ prompts consequently contain 2,062 tokens.

Evaluation is zero-shot, with no chat template or context repetition. Greedy cached decoding generates at most 48 tokens, stopping at a newline or EOS, in BF16 batches of up to eight examples of equal input length. An answer is correct if an accepted reference appears in the generated text as a case-insensitive substring. We report the unweighted mean of the six dataset accuracies.

\textbf{Search-scale reporting.}
Tables~\ref{tab:small-search-tasks} and~\ref{tab:small-search-niah} report the per-dataset measurements that determine the search utilities in Table~\ref{tab:search}. Tables~\ref{tab:small-full-tasks} and~\ref{tab:small-full-niah} give the corresponding full evaluations.

\begin{table}[!htbp]
\centering
\caption{\textbf{Search-scale LM and CSR feedback.} CSR uses 256 fixed examples per task; accuracies are percentages. These reduced-suite scores determine the utilities in Table~\ref{tab:search}, rather than its full-evaluation columns.}
\label{tab:small-search-tasks}
\small
\setlength{\tabcolsep}{5pt}
\begin{tabular}{@{}lcccc@{}}
\toprule \midrule
\textbf{Dataset / metric} & \textbf{GDN} & \textbf{KDA} & \textbf{GDN2} & \textbf{CFDN}\\
\midrule
WikiText-2 token PPL & 14.39 & 14.23 & 14.32 & 13.99\\
LAMBADA token PPL & 11.39 & 10.61 & 10.40 & 10.13\\
PIQA & 63.28 & 61.33 & 63.28 & 63.28\\
HellaSwag (normalized) & 34.77 & 34.38 & 34.38 & 36.33\\
WinoGrande & 53.91 & 52.73 & 50.78 & 53.12\\
ARC-Easy & 50.39 & 55.08 & 52.73 & 53.12\\
ARC-Challenge & 19.53 & 20.70 & 22.27 & 18.36\\
OpenBookQA & 18.36 & 21.88 & 22.66 & 22.66\\
Social IQa & 39.45 & 42.58 & 44.14 & 44.53\\
BoolQ & 64.06 & 60.55 & 65.62 & 65.62\\
\midrule
CSR mean & 42.97 & 43.65 & 44.48 & 44.63\\
\midrule \bottomrule
\end{tabular}
\end{table}

\begin{table}[!htbp]
\centering
\caption{\textbf{Search retrieval feedback (\%).} Each cell uses 64 fixed examples and exact teacher-forced continuation scoring, giving 768 requests overall.}
\label{tab:small-search-niah}
\small
\setlength{\tabcolsep}{4pt}
\begin{tabular}{@{}lccccc@{}}
\toprule \midrule
\textbf{Task} & \textbf{Context} & \textbf{GDN} & \textbf{KDA} & \textbf{GDN2} & \textbf{CFDN}\\
\midrule
S-NIAH-1 & 1K & 82.81 & 56.25 & 68.75 & 100.00\\
 & 2K & 42.19 & 42.19 & 50.00 & 98.44\\
 & 4K & 21.88 & 9.38 & 25.00 & 98.44\\
S-NIAH-2 & 1K & 42.19 & 62.50 & 82.81 & 96.88\\
 & 2K & 10.94 & 39.06 & 40.62 & 92.19\\
 & 4K & 1.56 & 7.81 & 15.62 & 25.00\\
S-NIAH-3 & 1K & 1.56 & 3.12 & 0.00 & 10.94\\
 & 2K & 0.00 & 4.69 & 0.00 & 10.94\\
 & 4K & 0.00 & 0.00 & 0.00 & 4.69\\
MK-NIAH-1 & 1K & 15.62 & 18.75 & 25.00 & 21.88\\
 & 2K & 9.38 & 15.62 & 26.56 & 26.56\\
 & 4K & 10.94 & 12.50 & 12.50 & 14.06\\
\midrule
Mean &  & 19.92 & 22.66 & 28.91 & 50.00\\
\midrule \bottomrule
\end{tabular}
\end{table}

\begin{table}[!htbp]
\centering
\caption{\textbf{Complete 150M LM and CSR results} for Tables~\ref{tab:search} and~\ref{tab:objective}. The last column is the strongest model selected by LM-only search. Accuracies are percentages; LAMBADA accuracy is excluded from the eight-task CSR mean.}
\label{tab:small-full-tasks}
\small
\setlength{\tabcolsep}{4pt}
\begin{tabular}{@{}lccccc@{}}
\toprule \midrule
\textbf{Dataset / metric} & \textbf{GDN} & \textbf{KDA} & \textbf{GDN2} & \textbf{CFDN} & \textbf{LM only}\\
\midrule
WikiText-2 word PPL & 35.15 & 34.63 & 34.93 & 33.84 & 34.59\\
LAMBADA PPL & 64.03 & 56.64 & 54.74 & 52.33 & 46.40\\
LAMBADA accuracy & 26.28 & 27.15 & 27.61 & 28.35 & 29.92\\
PIQA & 63.60 & 63.76 & 64.53 & 64.64 & 63.17\\
HellaSwag (normalized) & 34.25 & 34.94 & 34.57 & 35.22 & 34.47\\
WinoGrande & 51.62 & 51.07 & 53.04 & 51.22 & 52.17\\
ARC-Easy & 52.74 & 53.32 & 53.16 & 54.04 & 53.24\\
ARC-Challenge & 22.70 & 21.67 & 23.46 & 22.27 & 21.59\\
OpenBookQA & 19.00 & 20.00 & 20.00 & 20.40 & 20.80\\
Social IQa & 37.77 & 37.41 & 37.87 & 36.49 & 37.36\\
BoolQ & 61.41 & 53.30 & 62.17 & 62.02 & 55.23\\
\midrule
CSR mean & 42.88 & 41.94 & 43.60 & 43.29 & 42.25\\
\midrule \bottomrule
\end{tabular}
\end{table}

\begin{table}[!htbp]
\centering
\caption{\textbf{Complete 150M generated-answer NIAH results (\%)} for Tables~\ref{tab:search} and~\ref{tab:objective}. Each cell contains 500 frozen requests; the mean weights all 14 cells equally.}
\label{tab:small-full-niah}
\small
\setlength{\tabcolsep}{4pt}
\begin{tabular}{@{}lcccccc@{}}
\toprule \midrule
\textbf{Task} & \textbf{Context} & \textbf{GDN} & \textbf{KDA} & \textbf{GDN2} & \textbf{CFDN} & \textbf{LM only}\\
\midrule
S-NIAH-1 & 1K & 88.40 & 63.00 & 87.60 & 89.40 & 94.20\\
 & 2K & 42.40 & 38.60 & 49.80 & 33.20 & 73.40\\
 & 4K & 14.20 & 10.60 & 19.20 & 1.00 & 33.40\\
 & 8K & 6.40 & 3.60 & 8.40 & 0.40 & 12.40\\
S-NIAH-2 & 1K & 39.20 & 69.20 & 82.40 & 89.60 & 56.40\\
 & 2K & 5.00 & 42.00 & 45.60 & 81.00 & 39.20\\
 & 4K & 2.80 & 11.40 & 9.40 & 33.60 & 4.40\\
 & 8K & 0.00 & 7.40 & 3.80 & 7.40 & 4.00\\
S-NIAH-3 & 1K & 0.80 & 3.60 & 1.20 & 12.00 & 1.60\\
 & 2K & 0.20 & 1.20 & 0.20 & 6.60 & 0.40\\
 & 4K & 0.00 & 0.60 & 0.00 & 5.00 & 0.00\\
MK-NIAH-1 & 1K & 12.00 & 18.40 & 21.80 & 19.40 & 19.60\\
 & 2K & 16.60 & 19.80 & 17.00 & 19.40 & 20.00\\
 & 4K & 9.20 & 19.00 & 14.60 & 18.60 & 15.20\\
\midrule
Mean &  & 16.94 & 22.03 & 25.79 & 29.76 & 26.73\\
\midrule \bottomrule
\end{tabular}
\end{table}

\begin{table}[!htbp]
\centering
\caption{\textbf{Complete 1.3B LM and CSR results} for Table~\ref{tab:large}. Accuracies are percentages. LAMBADA accuracy is reported separately and excluded from the eight-task CSR mean.}
\label{tab:full-tasks}
\small
\setlength{\tabcolsep}{4pt}
\begin{tabular}{@{}lcccc@{}}
\toprule \midrule
\textbf{Dataset / metric} & \textbf{GDN} & \textbf{KDA} & \textbf{GDN2} & \textbf{CFDN}\\
\midrule
WikiText-2 word PPL & 16.27 & 16.05 & 16.05 & 15.74\\
LAMBADA PPL & 12.04 & 11.66 & 11.22 & 11.04\\
LAMBADA accuracy & 46.61 & 47.76 & 48.81 & 48.77\\
PIQA & 73.01 & 73.01 & 72.69 & 73.78\\
HellaSwag (normalized) & 57.47 & 57.63 & 57.78 & 58.53\\
WinoGrande & 57.46 & 58.80 & 57.77 & 58.48\\
ARC-Easy & 71.42 & 71.89 & 70.03 & 73.15\\
ARC-Challenge & 37.63 & 38.14 & 36.69 & 37.12\\
OpenBookQA & 29.20 & 29.60 & 28.80 & 30.80\\
Social IQa & 40.99 & 42.02 & 41.15 & 41.56\\
BoolQ & 62.66 & 60.09 & 56.30 & 62.81\\
\midrule
CSR mean & 53.73 & 53.90 & 52.65 & 54.53\\
\midrule \bottomrule
\end{tabular}
\end{table}

\begin{table}[!htbp]
\centering
\caption{\textbf{Complete generated-answer NIAH results at 1.3B (\%).} Each of the 14 cells contains 500 frozen requests shared across models. These scores use answer generation, rather than the search-time teacher-forced proxy.}
\label{tab:full-niah}
\small
\setlength{\tabcolsep}{4pt}
\begin{tabular}{@{}lccccc@{}}
\toprule \midrule
\textbf{Task} & \textbf{Context} & \textbf{GDN} & \textbf{KDA} & \textbf{GDN2} & \textbf{CFDN}\\
\midrule
S-NIAH-1 & 1K & 100.00 & 100.00 & 100.00 & 100.00\\
 & 2K & 88.20 & 100.00 & 100.00 & 100.00\\
 & 4K & 40.00 & 100.00 & 100.00 & 99.60\\
 & 8K & 18.60 & 99.60 & 98.60 & 94.20\\
S-NIAH-2 & 1K & 100.00 & 98.40 & 99.80 & 100.00\\
 & 2K & 100.00 & 100.00 & 99.80 & 100.00\\
 & 4K & 85.60 & 75.80 & 89.20 & 74.20\\
 & 8K & 13.80 & 37.60 & 34.60 & 21.20\\
S-NIAH-3 & 1K & 56.40 & 1.60 & 31.20 & 32.40\\
 & 2K & 59.80 & 1.60 & 54.80 & 74.00\\
 & 4K & 14.00 & 1.60 & 18.80 & 50.40\\
MK-NIAH-1 & 1K & 41.40 & 42.80 & 45.40 & 35.60\\
 & 2K & 28.00 & 35.20 & 36.20 & 31.00\\
 & 4K & 24.00 & 33.40 & 27.60 & 29.40\\
\midrule
Mean &  & 54.99 & 59.11 & 66.86 & 67.29\\
\midrule \bottomrule
\end{tabular}
\end{table}

\begin{table}[!htbp]
\centering
\caption{\textbf{Full real-world retrieval results at 1.3B (\%).} The mean weights the six datasets equally. Each model uses the same checkpoint as the LM, CSR, and NIAH results in Table~\ref{tab:large}. Best scores are bolded.}
\label{tab:full-real-retrieval}
\small
\setlength{\tabcolsep}{5pt}
\begin{tabular}{@{}lcccc@{}}
\toprule \midrule
\textbf{Dataset} & \textbf{GDN} & \textbf{KDA} & \textbf{GDN2} & \textbf{CFDN}\\
\midrule
SWDE & \textbf{31.96} & 26.62 & 26.15 & 27.55\\
SQuAD & 37.42 & 36.21 & \textbf{38.97} & 35.98\\
FDA & 27.16 & 22.80 & 25.98 & \textbf{31.06}\\
TriviaQA & 60.31 & \textbf{63.68} & 60.78 & 61.08\\
Natural Questions & 21.79 & \textbf{23.03} & 22.27 & 21.44\\
DROP & 20.41 & 22.47 & 20.99 & \textbf{23.33}\\
\midrule
Mean & 33.17 & 32.47 & 32.52 & \textbf{33.41}\\
\midrule \bottomrule
\end{tabular}
\end{table}

\section{Organizational settings and analysis protocol}
\label{app:analysis}

\subsection{Scientific profiles and researcher initialization}
\label{app:profiles}

The team uses 16 fixed disciplinary profiles. Each is a self-contained document of approximately 2,000--2,300 words, combining a field's mathematical foundations, design judgments, diagnostic practices, and primary references. Table~\ref{tab:profiles} summarizes the different research questions these documents foreground. Appendix~\ref{app:profile-exact} reproduces the complete Bayesian-filtering profile supplied to one researcher. Only the focal linear attention profile is written specifically about the target architecture family.

At initialization, each researcher receives one profile, the common problem statement, and access to shared references covering the baseline, prior work, and experimental constraints. Its conversation, private notebook, and earlier proposal dossiers preserve individual experience as the search proceeds. At each new assignment, the researcher additionally receives a summary of previous results, ongoing investigations, and consolidated shared lessons, with access to the underlying records. The fixed profile therefore supplies an initial technical perspective, while private and shared evidence can change the researcher's subsequent judgments.

\begin{table}[htbp]
\centering
\caption{\textbf{The 16 scientific perspectives.} Each row summarizes one profile's organizing question; it is not a prescribed candidate mechanism or an exclusive research territory.}
\label{tab:profiles}
\small
\setlength{\tabcolsep}{4pt}
\begin{tabular}{@{}>{\raggedright\arraybackslash}p{4.1cm}>{\raggedright\arraybackslash}p{9.1cm}@{}}
\toprule \midrule
\textbf{Scientific perspective} & \textbf{Organizing question or distinction}\\
\midrule
Conditional routing and adaptive computation & When should state change, which path should act, and when is further computation worthwhile?\\
\addlinespace[3pt]
Multiscale and hierarchical recurrence & How should clocks, temporal boundaries, and communication represent phenomena at different rates?\\
\addlinespace[3pt]
Stable, switched, and non-normal dynamics & Which distinctions survive input-driven transitions, and when do perturbations amplify?\\
\addlinespace[3pt]
Multiplicative operators and automata & What computations can input-selected operators express through composition?\\
\addlinespace[3pt]
Parallel and I/O-aware recurrent design & Which algebraic structure permits efficient execution once state traffic and differentiation are counted?\\
\addlinespace[3pt]
Streaming sketches and mergeable summaries & Which queries must bounded state support, and how does error compose across updates and merges?\\
\addlinespace[3pt]
Cache policies and memory hierarchies & What deserves residency, when should it move, and when is a stored copy no longer useful?\\
\addlinespace[3pt]
Differentiable computers and neural data structures & Do learned addressing, allocation, and traversal retain their meaning under repeated operations?\\
\addlinespace[3pt]
Predictive representations and information bottlenecks & Which distinctions in history matter for future predictions, and which can be discarded?\\
\addlinespace[3pt]
Bayesian filtering and regime change & How should uncertainty determine correction, and how can an outlier be distinguished from a change of regime?\\
\addlinespace[3pt]
Continual memory and consolidation & How can new information be acquired without overwriting useful prior structure?\\
\addlinespace[3pt]
Associative and sparse adaptive memory & How do retrieval, residual competition, and allocation interact under finite capacity?\\
\addlinespace[3pt]
Vector-symbolic structured memory & Which binding, superposition, and cleanup operations preserve compositional meaning?\\
\addlinespace[3pt]
Tensor, spectral, and low-rank operator memory & What state rank preserves distinctions needed by future computations?\\
\addlinespace[3pt]
Online recurrent learning and eligibility traces & How can delayed credit be assigned with bounded storage and causal updates?\\
\addlinespace[3pt]
Linear attention and recurrent matrix memory & How should a fixed-size matrix state read, write, correct, and forget while supporting parallel training?\\
\midrule \bottomrule
\end{tabular}
\end{table}

\subsection{Research protocol and analysis details}

\textbf{Proposal and review details.}
Each proposal specifies its hypothesis, expected benefit, and a falsifying outcome, together with state equations, parameterization, a parallelizable training formulation, and an analytical compute breakdown. For lineage analysis, it identifies one primary scientific precursor and may cite other influences; this attribution does not require retaining the precursor's architecture. Technical review precedes contextual review. Both reviewers are blind to the author's identity and the other review. The technical reviewer assesses the proposal without the shared research history, while the contextual reviewer uses that history to assess duplication and research value. The original researcher then resolves correctable issues into the revised specification.

\textbf{Memory and information access.}
The persistent condition preserves the originating researcher's conversation and notebook across iterations. The fresh condition preserves them only within one iteration, including proposal revision and terminal reflection. It still receives shared proposals, results, published lessons, and the shared result summary. Reviewers and implementers use task-specific sessions in both conditions. The evidence available at reflection includes reviews, source code, records of implementation attempts, training logs, and evaluation artifacts. A director consolidates shared lessons after every ten completed reflections; it does not select the next researcher. Agent tools include web search and a numerical Python environment. Model source and GPU execution belong to the implementation role, rather than the proposing researcher.

\textbf{Allocation and displayed ranks.}
The adaptive policy uses $\gamma=2^{-1/32}$ and $c=0.25$ in Eq.~(\ref{eq:ucb}). An observation's weight therefore halves after 32 completed reflections, without changing the recorded model utility or the reward originally assigned to its researcher. Researchers with unresolved investigations are ineligible. Uniform selection repeats the first-round researcher order and waits if the next researcher is busy; it does not skip ahead according to completion time or reward. In the shared-prefix experiment, this gives every researcher four new assignments. Ties in the adaptive comparison use a stable researcher ordering common to the continuation arms.

The shared result summary displays numeric utility for every complete comparable endpoint. It shows overall top-five ranks only for candidates whose regressions relative to GDN2 do not exceed 0.01 NLL, 0.2 CSR percentage points, or 2 retrieval percentage points. A candidate outside these tolerances retains its numerical utility and a marker naming the regressing target. This display rule does not filter scheduler rewards or the highest-utility reference used for researcher credit (Section~\ref{sec:progress}).

\textbf{Common first-round evidence.}
The fresh-memory and uniform-selection runs import the main run's first 16 investigations. Their proposals, reviews, specifications, implementation evidence, and outcomes are identical at the branch. Uniform selection retains the reconstructed private conversations and notes; fresh memory creates a new private researcher session for every subsequent iteration. Both can read the common shared evidence, including published reflections and lessons, and conduct 64 new iterations.

\textbf{Recorded parent relationships.}
Table~\ref{tab:process} summarizes relationships declared in proposals. We count a parent--child pair only when the parent is an earlier candidate, excluding GDN2. A pair is eligible for the utility comparison when both models have complete evaluations; an improvement means strictly higher child utility. For cross-researcher links, we count proposals whose declared candidate parent belongs to a different researcher, including proposals without a completed evaluation. The denominators therefore differ. These are counts of declared relationships and measured outcomes, rather than annotations of mechanism novelty or proof of knowledge transfer.

\begin{table}[!htbp]
\centering
\caption{\textbf{Completed evaluations and declared parent relationships.} Each row covers 80 iterations. A fraction reports the number satisfying the criterion over the number eligible. The fresh-memory and uniform-selection rows include the common first 16 iterations and their 11 completed evaluations.}
\label{tab:process}
\small
\setlength{\tabcolsep}{5pt}
\begin{tabular}{@{}lccc@{}}
\toprule \midrule
\textbf{Condition} & \textbf{\shortstack{Completed\\evaluations}} & \textbf{\shortstack{Higher utility\\than parent}} & \textbf{\shortstack{Cross-researcher\\parent links}}\\
\midrule
Main & 63 & 9/41 & 41/54\\
No background & 67 & 12/42 & 41/57\\
Fresh memory & 65 & 5/32 & 36/44\\
Uniform selection & 68 & 15/46 & 38/57\\
\midrule \bottomrule
\end{tabular}
\end{table}

\input{section/appendix_profile}

%% file: section/appendix_profile.tex
\subsection{An exact researcher profile}
\label{app:profile-exact}
The box below reproduces the complete Markdown file supplied to the Bayesian-filtering researcher, including its original mathematical notation and reference links.

\begin{tcolorbox}[
  breakable, enhanced,
  title={Bayesian Filtering, Innovations, and Regime Change (researcher profile)},
  colback=gray!5, colframe=gray!50,
  coltitle=black, fonttitle=\bfseries\small,
  left=2pt, right=2pt, top=2pt, bottom=2pt,
  boxrule=0.5pt
]
\begin{Verbatim}
# Bayesian Filtering, Innovations, and Regime Change

Sequential inference is the discipline of maintaining a calibrated belief about an evolving latent process. The state of a filter is not merely its current best estimate. It is a probability distribution—or a controlled approximation to one—whose uncertainty determines how new evidence should be interpreted.

Consider the state-space model

\[
x_t\sim p(x_t\mid x_{t-1}),\qquad
y_t\sim p(y_t\mid x_t).
\]

Bayesian filtering alternates prediction and correction:

\[
p(x_t\mid y_{1:t-1})
=\int p(x_t\mid x_{t-1})p(x_{t-1}\mid y_{1:t-1})\,dx_{t-1},
\]

\[
p(x_t\mid y_{1:t})
\propto p(y_t\mid x_t)p(x_t\mid y_{1:t-1}).
\]

This recursion is exact as an identity. All difficulty lies in representing and updating the posterior.

## Linear–Gaussian filtering and the meaning of innovation

For linear dynamics and Gaussian noise,

\[
x_t=F_tx_{t-1}+w_t,\quad w_t\sim\mathcal N(0,Q_t),
\]
\[
y_t=H_tx_t+v_t,\quad v_t\sim\mathcal N(0,R_t),
\]

the posterior remains Gaussian. After predicting mean \(m_t^-\) and covariance \(P_t^-\), the innovation and its covariance are

\[
\nu_t=y_t-H_tm_t^-,\qquad
C_t=H_tP_t^-H_t^\top+R_t.
\]

The Kalman gain \(K_t=P_t^-H_t^\top C_t^{-1}\) balances model confidence against measurement confidence, and the corrected mean is \(m_t=m_t^-+K_t\nu_t\).

The key quantity is the **normalized innovation**, not the raw residual. A residual of magnitude ten is unremarkable if predicted uncertainty is one hundred and alarming if uncertainty is one tenth. Innovation whiteness is also a diagnostic: persistent temporal structure in normalized innovations signals a misspecified dynamics model, observation model, or noise covariance.

Covariance is computational state, not decorative metadata. Discarding it forces a fixed response to evidence and prevents principled fusion of measurements with different reliability. Square-root filters propagate factors of \(P\) for numerical stability. Information-form filters propagate precision \(J=P^{-1}\) and information vector \(h=Jm\), making independent measurement contributions additive. Choice of form should follow sparsity and conditioning.

Consistency can be tested statistically. Under the assumed model, the normalized innovation squared

\[
\operatorname{NIS}_t=\nu_t^\top C_t^{-1}\nu_t
\]

has a chi-squared reference distribution with dimension equal to the observation dimension. A sequence of values that is systematically too large indicates overconfidence or mismatch; values that are systematically too small indicate excessive uncertainty. Temporal dependence must be checked separately with innovation-whiteness tests. Coverage and whiteness reveal failures that root-mean-square error cannot.

Even algebraically equivalent covariance updates differ in finite precision. The compact expression \(P_t=(I-K_tH_t)P_t^-\) can lose symmetry or positive semidefiniteness. The Joseph form,

\[
P_t=(I-K_tH_t)P_t^-(I-K_tH_t)^\top+K_tR_tK_t^\top,
\]

is more robust, and square-root implementations are preferable when conditioning is severe. Numerical representation is part of the inferential method because an invalid covariance implies invalid confidence.

## Beyond a single Gaussian

Extended Kalman filtering linearizes nonlinear dynamics locally and can become inconsistent when curvature is strong. Sigma-point methods propagate a designed set of points through the nonlinear map, improving moment approximation without explicit Jacobians but still reducing the posterior to a Gaussian.

Particle filters retain weighted samples:

\[
p(x_t\mid y_{1:t})\approx
\sum_{i=1}^{N}w_t^{(i)}\delta(x_t-x_t^{(i)}).
\]

They can represent multimodality and non-Gaussian tails, but importance weights degenerate over time. Resampling removes low-weight particles yet causes impoverishment and destroys ancestry diversity. The importance-sampling distribution is therefore central: sampling only from the transition prior wastes particles when the observation is informative. Differentiable or learned filters do not abolish these issues; they can simply hide miscalibration inside an expressive learned sampler.

Rao–Blackwellization is often preferable to brute-force particles. If part of the model is conditionally linear–Gaussian, integrate that part analytically and sample only the discrete or nonlinear component. Expert design exploits conditional structure before increasing particle count.

## Robustness is a modeling decision

An unexpectedly large innovation admits several explanations: the observation
may be corrupted, the latent state may have moved farther than expected, the
dynamics may have changed regime, or the posterior may simply have been
overconfident. A robust filter should not collapse these explanations into one
generic downweighting rule. Each implies a different posterior and a different
prediction for the next observation.

Heavy-tailed likelihoods, such as Student-(t) observation models, reduce the
leverage of isolated outliers. Mixture likelihoods can carry an explicit
contamination indicator. Adaptive process noise can accommodate gradual drift,
while switching dynamics or run-length uncertainty represent persistent
structural change. These mechanisms are not interchangeable. An overly heavy
tail can explain away a genuine change point forever; inflated process noise
can preserve tracking error at the cost of uninformative uncertainty; a reset
rule can mistake one sensor fault for a new regime.

I distinguish these cases through persistence and cross-channel structure. A
single-channel spike with immediate recovery supports contamination. Coherent
innovations across several measurements support latent motion or model error.
A sustained change in the distribution of normalized innovations supports a
regime hypothesis. Posterior predictive checks should reproduce not only
marginal residual magnitudes but also their temporal dependence, direction,
and coincidence across sensors.

Robustness must remain calibrated. A method that never issues a confident
posterior is difficult to falsify but not useful. I report coverage conditioned
on nominal periods, outliers, and transitions, together with detection delay
and false-alarm rate. This makes explicit whether apparent resilience comes
from correct hypothesis management or from indiscriminately broad intervals.

## Switching models and hypothesis management

Abruptly changing systems require uncertainty over both continuous state and discrete regime. If a mode \(z_t\) follows a Markov chain and each mode has its own dynamics, exact filtering yields a mixture whose number of components grows exponentially. The interacting multiple-model method controls this explosion by mixing, filtering within each mode, and merging hypotheses at every step.

This is an important general lesson: bounded inference requires a deliberate **hypothesis management policy**. Pruning, merging, and moment matching are not implementation afterthoughts; they determine which alternative explanations survive. A mode with low current probability may need preservation if it predicts a high-consequence future or is difficult to recover after deletion.

## Change points as uncertainty over run length

Bayesian online change-point detection introduces the run length \(r_t\), the number of observations since the most recent change. A hazard function \(H(r)\) specifies the prior probability of a change after a run of length \(r\). The posterior update combines two possibilities:

\[
p(r_t=0, y_{1:t})
=\sum_{r_{t-1}}
p(r_{t-1},y_{1:t-1})H(r_{t-1})
p(y_t\mid r_{t-1}),
\]

and growth transitions \(r_t=r_{t-1}+1\) weighted by \(1-H(r_{t-1})\).

This is more informative than a reset score. It maintains uncertainty over when the current regime began, and each run-length hypothesis carries its own sufficient statistics. The hazard encodes duration assumptions; a constant hazard implies a geometric regime duration and is often adopted without justification.

Exact run-length support grows by one at every observation. A bounded implementation must prune, merge, or aggregate old hypotheses, and this approximation determines the longest change history the detector can express. Keeping only the most probable run length turns posterior inference into brittle winner-take-all segmentation. A small tail mass spread over many long runs can matter even when no individual run is likely.

Outliers and change points are easily confused. A robust observation model should explain isolated anomalies without resetting the regime, whereas sustained predictive failure should transfer mass toward short run lengths. Heavy-tailed likelihoods, explicit outlier indicators, and hierarchical hazards help separate these phenomena.

## Event-triggered inference

In sensing and networked control, communication or correction can be triggered only when the innovation is informative enough to justify its cost. Crucially, silence can itself be evidence: if a sensor transmits only when a residual leaves a known region, no transmission tells the estimator that the residual remained inside that region. Treating skipped updates as missing at random loses information and biases uncertainty.

Event policies should therefore be designed jointly with the estimator. A threshold on \(\nu_t^\top C_t^{-1}\nu_t\) has a statistical interpretation, while a threshold on raw magnitude does not. Resource-aware filtering asks for an optimal trade between posterior risk, sensing cost, latency, and energy.

## Filtering, smoothing, and decision delay

Filtering estimates \(x_t\) from \(y_{1:t}\); smoothing permits later observations. Fixed-lag smoothing occupies a useful middle ground: it delays commitment by \(L\) steps and revises recent states using \(y_{t+1:t+L}\). This is often more valuable than making a much larger causal filter, especially near regime boundaries.

The correct comparison therefore fixes decision latency. Zero-lag detection, fixed-lag revision, and retrospective segmentation answer different questions. A change-point method should also state whether it estimates the first affected sample, the time at which evidence became sufficient, or the time at which an alarm was issued.

## Failure modes

- **Overconfidence:** underestimated \(Q\) or \(R\) produces brittle gains and poor coverage.
- **Model mismatch:** a precise posterior under the wrong model is worse than an honest broad posterior.
- **Innovation misuse:** large raw residuals are treated as changes without uncertainty normalization.
- **Particle collapse:** a nominally multimodal filter effectively contains one duplicated hypothesis.
- **Premature merging:** distinct modes with similar current means have incompatible futures.
- **Hazard misspecification:** expected regime duration silently dictates detection latency.
- **Numerical loss of positive definiteness:** naïve covariance updates accumulate roundoff error.
- **Retrospective leakage:** smoothing results are reported as if available to an online filter.

## An expert workflow for sequential inference

When a sequential estimator fails, I first decide which problem it is failing
to solve. State estimation asks where the process is within a trusted model.
Robust inference asks whether an observation is an outlier. Model selection
asks which dynamics explain the data. Change-point detection asks whether the
old regime has ended. These explanations can produce the same large residual,
but they imply different updates.

My operating sequence is:

1. **Write the belief state.** I state which variables are uncertain and which
   approximation represents their joint posterior: moments, information
   parameters, particles, regime mixtures, run lengths, or a factorization.
   A point estimate is acceptable only when uncertainty cannot affect any
   subsequent decision.
2. **Predict before observing.** The transition model produces a predictive
   distribution, not just a predicted mean. Its spread is the reference against
   which the next observation becomes ordinary or surprising.
3. **Correct by relative evidence.** I inspect likelihood ratios or normalized
   innovations. Raw residual magnitude has no context. With Gaussian
   observations, \(\nu_t^\top C_t^{-1}\nu_t\) is the natural scalar diagnostic;
   for non-Gaussian models, posterior predictive checks play the same role.
4. **Keep explanations alive when their futures differ.** A low-probability
   regime, run length, or particle family should not be merged merely because
   its present mean resembles the winner. Pruning is a resource decision about
   tail mass, recoverability, and future consequence. Conditional analytic
   structure should be Rao–Blackwellized before particles are multiplied.
5. **Interpret persistence.** One unlikely observation may call for a
   heavy-tailed likelihood or outlier indicator. A sequence of structured
   innovations suggests dynamics mismatch or regime change. The hazard model
   states how readily persistent failure becomes a reset; it should never be
   an invisible constant.
6. **Choose commitment latency.** Filtering, fixed-lag smoothing, and
   retrospective segmentation permit different evidence. I compare methods at
   equal delay and distinguish the first affected time, the time evidence
   became sufficient, and the alarm time.
7. **Audit numerical and statistical consistency.** I preserve symmetry and
   positive semidefiniteness with square-root or Joseph-form updates when
   needed. I check posterior coverage, normalized-innovation distributions,
   innovation whiteness, effective particle diversity, and recovery after
   mismatch—not point error alone.

Missingness and silence require an observation model too. If a sensor reports
only outside a threshold region, no report truncates the residual distribution
and is informative. If observations are missing for external reasons, treating
the gap as evidence is wrong. Sensing, communication, and inference policies
must therefore be specified jointly whenever reporting depends on latent
state or innovation.

To falsify an estimator, I vary noise scale, inject isolated heavy-tailed
outliers, switch among regimes with similar current means but different
dynamics, alter regime duration, and withhold informative observations. I
check whether uncertainty broadens before errors grow, whether alternative
hypotheses survive long enough to recover, and whether apparent calibration
comes from inflating \(Q\) or \(R\) to conceal model-form error. Learned
dynamics, observation models, and samplers are useful only if their flexibility
does not make these checks vacuous; epistemic uncertainty and irreducible
observation noise should not be collapsed into one convenient variance.

The unresolved resource problem is hypothesis management. Exact belief support
usually grows, while hard winner-take-all truncation is brittle. Pruning,
merging, moment matching, and learned sampling mechanisms should be judged by
which future distinctions they erase and how their lost tail mass is
accounted for.

The defining principle is that an observation has no fixed importance. Its
importance is the likelihood ratio it induces relative to the current
predictive belief.

## Selected primary references

- Kalman, [“A New Approach to Linear Filtering and Prediction Problems”](https://doi.org/10.1115/1.3662552).
- Gordon, Salmond, and Smith, [“Novel Approach to Nonlinear/Non-Gaussian Bayesian State Estimation”](https://people.bordeaux.inria.fr/pierre.delmoral/gordon-salmond-smith-1993.pdf).
- Blom and Bar-Shalom, [“The Interacting Multiple Model Algorithm for Systems with Markovian Switching Coefficients”](https://doi.org/10.1109/9.1299).
- Adams and MacKay, [“Bayesian Online Changepoint Detection”](https://arxiv.org/abs/0710.3742).
- Jonschkowski et al., [“Differentiable Particle Filters”](https://arxiv.org/abs/1805.11122).
\end{Verbatim}
\end{tcolorbox}

%% file: section/appendix_architecture.tex
\section{Architecture details}
\label{app:architecture-all}

\subsection{The discovered context-first recurrence}
\label{app:architecture}

We describe Context-First DeltaNet (CFDN), introduced in Sections~\ref{sec:discovery} and~\ref{sec:lineage}. GDN2 stores information in a recurrent matrix and retrieves it through a query-dependent read. CFDN adds a recurrent context vector that contributes the same additive term to every query. This gives the memory two ways to represent information: a shared context contribution and an address-dependent contribution from the matrix.

The central idea is how these two memories learn from a prediction error. At each token, they jointly predict the gated value. CFDN corrects the context vector first, then updates the matrix using the error that remains. This gives shared context priority and encourages the matrix to represent corrections beyond what context already explains.

\noindent\textbf{Correct context before fitting the remaining error.}
For one head, let $\mathbf{S}_t\in\mathbb{R}^{d_k\times d_v}$ be the matrix memory and $\mathbf{m}_t\in\mathbb{R}^{d_v}$ the context vector, both initialized to zero. Let $\mathbf{q}_t,\mathbf{k}_t\in\mathbb{R}^{d_k}$ be the normalized query and key, $\mathbf{v}_t\in\mathbb{R}^{d_v}$ the value, and $\mathbf{b}_t\in(0,1)^{d_k}$ and $\mathbf{w}_t\in(0,1)^{d_v}$ the sigmoid erase and write gates.
As in GDN2, a channel-wise log-decay $\mathbf{g}_t$ first acts on the matrix:
\begin{equation}
 \bar{\mathbf{S}}_t
 =\operatorname{Diag}(\exp\mathbf{g}_t)\mathbf{S}_{t-1}.
\end{equation}
The context vector has no passive decay.
Denote the matrix's write and prediction addresses by $\mathbf{a}_t$ and $\mathbf{e}_t$, the context weight by $c$, and the correction normalizer by $\kappa_t$. Their construction is given below.
The context contribution to the predictor uses the mean erase gate, $\beta_t=d_k^{-1}\sum_{j=1}^{d_k}b_{t,j}$. The update has four steps:
\begin{equation}
\begin{aligned}
 \boldsymbol{\delta}_t
 &=\underbrace{\mathbf{w}_t\odot\mathbf{v}_t}_{\text{gated value}}
   -\underbrace{\left(
      \bar{\mathbf{S}}_t^\top\mathbf{e}_t
      +c\beta_t\mathbf{m}_{t-1}
    \right)}_{\text{joint prediction}},\\
 \mathbf{m}_t
 &=\mathbf{m}_{t-1}
   +\frac{c}{\kappa_t}\boldsymbol{\delta}_t,\\
 \boldsymbol{\delta}_t^{\mathrm{rem}}
 &=\mathbf{w}_t\odot\mathbf{v}_t
   -\bar{\mathbf{S}}_t^\top\mathbf{e}_t
   -c\beta_t\mathbf{m}_t,\\
 \mathbf{S}_t
 &=\bar{\mathbf{S}}_t
   +\frac{\mathbf{a}_t}{\kappa_t}
    \left(\boldsymbol{\delta}_t^{\mathrm{rem}}\right)^\top.
\end{aligned}
\label{eq:context-first}
\end{equation}
The first line measures what the two memories fail to predict together. The second assigns part of that error to shared context. The third recomputes the error after this context correction, while keeping the matrix fixed. Only then does the fourth line update the matrix.

Equivalently, the error passed to the matrix is
\begin{equation}
 \boldsymbol{\delta}_t^{\mathrm{rem}}
 =\left(1-\frac{c^2\beta_t}{\kappa_t}\right)
  \boldsymbol{\delta}_t.
\end{equation}
Thus, the two corrections can be combined algebraically into a single rank-one update of the augmented memory, retaining compatibility with parallel chunkwise training.

\noindent\textbf{Choose what the shared context represents.}
CFDN learns a reference address, or origin, $\boldsymbol{\mu}$ for each head. The matrix reads and writes relative to this origin. It also learns a scalar $\rho$ that controls the relative weighting of the matrix and context coordinates:
\begin{equation}
 \rho=\frac{\sigma(\theta)}{8},\qquad
 s=\sqrt{1-\rho},\qquad
 c=\sqrt{\rho},\qquad
 \boldsymbol{\mu}
 =\frac{\mathbf{u}}{8\sqrt{1+\lVert\mathbf{u}\rVert^2}},
\end{equation}
where $\theta\in\mathbb{R}$ and $\mathbf{u}\in\mathbb{R}^{d_k}$ are learned parameters. Unlike the recurrent context vector, these parameters are fixed across tokens.
The matrix addresses are
\begin{equation}
 \mathbf{a}_t=s(\mathbf{k}_t-\boldsymbol{\mu}),\qquad
 \mathbf{e}_t=\mathbf{b}_t\odot\mathbf{a}_t.
\end{equation}
The write address determines where the matrix stores a correction; the prediction address additionally applies GDN2's channel-wise erase gate. Translation changes the address energy, so CFDN scales both corrections using
\begin{equation}
 \kappa_t
 =1+(1-\rho)\left(
   \lVert\boldsymbol{\mu}\rVert^2
   -2\mathbf{k}_t^\top\boldsymbol{\mu}
 \right).
\end{equation}
This compensates for the energy change caused by translation and gives $\kappa_t=1$ when $\boldsymbol{\mu}=\mathbf{0}$.

After updating both memories, the raw readout is
\begin{equation}
 \mathbf{o}_t
 =\frac{
   s\mathbf{S}_t^\top(\mathbf{q}_t-\boldsymbol{\mu})
   +c\mathbf{m}_t
 }{\sqrt{d_k}}.
\end{equation}
The context term contributes equally to every query, while the matrix term varies with the query's displacement from the learned origin. At $\mathbf{q}_t=\boldsymbol{\mu}$, the matrix contribution vanishes, so passive matrix decay alone leaves this readout unchanged. Subsequent corrections can still change its readout through $\mathbf{m}_t$.

We initialize $\mathbf{u}=\mathbf{0}$ and $\theta=-\log 15$, giving $\rho=1/128$. The parameterization ensures $\lVert\boldsymbol{\mu}\rVert<1/8$ and $0<\rho<1/8$.
Our models use $d_k=d_v=128$ and retain GDN2's surrounding components: the query/key/value projections and short convolutions, independent erase/write projections, and per-head RMS normalization followed by sigmoid output gating and the output projection suggested by~\cite{kimiteam2025kimilinearexpressiveefficient}.